\documentclass[10pt,a4paper]{article}
\usepackage[utf8]{inputenc}
\usepackage[english]{babel}
\usepackage{amsmath}
\usepackage{amsfonts}
\usepackage{amssymb}
\usepackage{standalone}
\usepackage{tikz}
\usepackage{pgfplots}
\usetikzlibrary{matrix,positioning}
\usetikzlibrary{decorations.pathreplacing,patterns}
\usepackage{graphicx} 			
\usepackage{authblk}
\usepackage[
	colorlinks=true,
	urlcolor=blue,
	linkcolor=green
]{hyperref}	
\author{Marius Jahrens \qquad Thomas Martinetz }
\affil{Institute for Neuro- and Bioinformatics\\ University of Lübeck\\
\texttt{jahrens@inb.uni-luebeck.de}\\
\texttt{martinetz@inb.uni-luebeck.de}}
\title{Multi-layer Relation Networks}
\date{}

\begin{document}
\maketitle
\begin{abstract}
Relational Networks (RN) as introduced by Santoro et al. (2017) \cite{santoro17} have demonstrated strong relational reasoning capabilities with a rather shallow architecture. Its single-layer design, however, only considers pairs of information objects, making it unsuitable for problems requiring reasoning across a higher number of facts. To overcome this limitation, we propose a multi-layer relation network architecture which enables successive refinements of relational information through multiple layers. We show that the increased depth allows for more complex relational reasoning by applying it to the bAbI 20 QA dataset, solving all 20 tasks with joint training and surpassing the state-of-the-art results. 


\end{abstract}

\section{Introduction}
While Neural Networks have been very successful in classification and regression, their ability to cope with relational tasks has remained rather limited. In order to overcome this limitation, Neural Networks need a way to perform reasoning on different aspects of an input signal, so that more complex relations can be deducted through logical chains. 

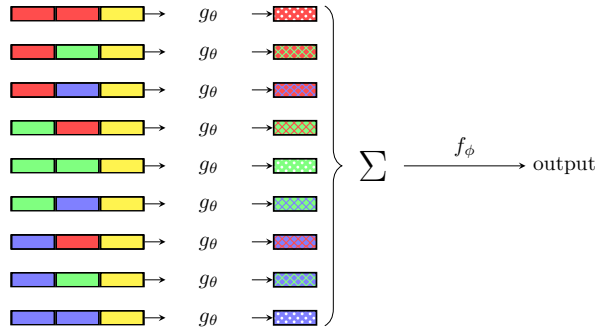
\begin{figure}
	\centering
	\tikzset{
block/.style={
	rectangle,
	draw,
	text centered
},
block2/.style={
	rectangle,
	draw,
	fill=gray,
	thick
},
block3/.style={
	rectangle,
	draw,
	thick,
	color=black,
	minimum width=2em,
	anchor=west
},
block4/.style={
	thick,
	rectangle,
	draw,
	minimum width=2em
},
block5/.style={
	block3,
	pattern=north west lines,
	pattern color=white
}
}

\begin{tikzpicture}[scale=1.0]
\matrix (m2) [matrix of nodes,row sep=0.5em,column sep=1em,minimum width=4em,] {
	|[draw,minimum width=6em+2.4pt,path picture={
		\node at (current bounding box.west) [block3,fill=red!70] (o1) {};
		\node at (o1.east) [block3,fill=red!70] (o2) {};
		\node at (o2.east) [block3,fill=yellow!80] (q) {};
	}]| {} & 
	|[yshift=-2pt]| {$g_\theta$} & 
	|[preaction={fill, red!70},pattern=crosshatch dots,pattern color=white,block4]| {} \\
	|[draw,minimum width=6em+2.4pt,path picture={
		\node at (current bounding box.west) [block3,fill=red!70] (o1) {};
		\node at (o1.east) [block3,fill=green!50] (o2) {};
		\node at (o2.east) [block3,fill=yellow!80] (q) {};
	}]| {} & 
	|[yshift=-2pt]| {$g_\theta$} & 
	|[preaction={fill, red!70},pattern=crosshatch,pattern color=green!50,block4]| {} \\
	|[draw,minimum width=6em+2.4pt,path picture={
		\node at (current bounding box.west) [block3,fill=red!70] (o1) {};
		\node at (o1.east) [block3,fill=blue!50] (o2) {};
		\node at (o2.east) [block3,fill=yellow!80] (q) {};
	}]| {} & 
	|[yshift=-2pt]| {$g_\theta$} & 
	|[preaction={fill, red!70},pattern=crosshatch,pattern color=blue!50,block4]| {} \\
	|[draw,minimum width=6em+2.4pt,path picture={
		\node at (current bounding box.west) [block3,fill=green!50] (o1) {};
		\node at (o1.east) [block3,fill=red!70] (o2) {};
		\node at (o2.east) [block3,fill=yellow!80] (q) {};
	}]| {} & 
	|[yshift=-2pt]| {$g_\theta$} & 
	|[preaction={fill, green!50},pattern=crosshatch,pattern color=red!70,block4]| {} \\
	|[draw,minimum width=6em+2.4pt,path picture={
		\node at (current bounding box.west) [block3,fill=green!50] (o1) {};
		\node at (o1.east) [block3,fill=green!50] (o2) {};
		\node at (o2.east) [block3,fill=yellow!80] (q) {};
	}]| {} & 
	|[yshift=-2pt]| {$g_\theta$} & 
	|[preaction={fill, green!50},pattern=crosshatch dots,pattern color=white,block4]| {} \\
	|[draw,minimum width=6em+2.4pt,path picture={
		\node at (current bounding box.west) [block3,fill=green!50] (o1) {};
		\node at (o1.east) [block3,fill=blue!50] (o2) {};
		\node at (o2.east) [block3,fill=yellow!80] (q) {};
	}]| {} & 
	|[yshift=-2pt]| {$g_\theta$} & 
	|[preaction={fill, green!50},pattern=crosshatch,pattern color=blue!50,block4]| {} \\
	|[draw,minimum width=6em+2.4pt,path picture={
		\node at (current bounding box.west) [block3,fill=blue!50] (o1) {};
		\node at (o1.east) [block3,fill=red!70] (o2) {};
		\node at (o2.east) [block3,fill=yellow!80] (q) {};
	}]| {} & 
	|[yshift=-2pt]| {$g_\theta$} & 
	|[preaction={fill, blue!50},pattern=crosshatch,pattern color=red!70,block4]| {} \\
	|[draw,minimum width=6em+2.4pt,path picture={
		\node at (current bounding box.west) [block3,fill=blue!50] (o1) {};
		\node at (o1.east) [block3,fill=green!50] (o2) {};
		\node at (o2.east) [block3,fill=yellow!80] (q) {};
	}]| {} & 
	|[yshift=-2pt]| {$g_\theta$} & 
	|[preaction={fill, blue!50},pattern=crosshatch,pattern color=green!50,block4]| {} \\
	|[draw,minimum width=6em+2.4pt,path picture={
		\node at (current bounding box.west) [block3,fill=blue!50] (o1) {};
		\node at (o1.east) [block3,fill=blue!50] (o2) {};
		\node at (o2.east) [block3,fill=yellow!80] (q) {};
	}]| {} & 
	|[yshift=-2pt]| {$g_\theta$} & 
	|[preaction={fill, blue!50},pattern=crosshatch dots,pattern color=white,block4]| {} \\
};

\path[-stealth]
	(m2-1-1) edge node {} (m2-1-2.west |- m2-1-1)
	(m2-1-2.east |- m2-1-3) edge node {} (m2-1-3)
	(m2-2-1) edge node {} (m2-2-2.west |- m2-2-1)
	(m2-2-2.east |- m2-2-3) edge node {} (m2-2-3)
	(m2-3-1) edge node {} (m2-3-2.west |- m2-3-1)
	(m2-3-2.east |- m2-3-3) edge node {} (m2-3-3)
	(m2-4-1) edge node {} (m2-4-2.west |- m2-4-1)
	(m2-4-2.east |- m2-4-3) edge node {} (m2-4-3)
	(m2-5-1) edge node {} (m2-5-2.west |- m2-5-1)
	(m2-5-2.east |- m2-5-3) edge node {} (m2-5-3)
	(m2-6-1) edge node {} (m2-6-2.west |- m2-6-1)
	(m2-6-2.east |- m2-6-3) edge node {} (m2-6-3)
	(m2-7-1) edge node {} (m2-7-2.west |- m2-7-1)
	(m2-7-2.east |- m2-7-3) edge node {} (m2-7-3)
	(m2-8-1) edge node {} (m2-8-2.west |- m2-8-1)
	(m2-8-2.east |- m2-8-3) edge node {} (m2-8-3)
	(m2-9-1) edge node {} (m2-9-2.west |- m2-9-1)
	(m2-9-2.east |- m2-9-3) edge node {} (m2-9-3);


\draw [decorate,decoration={brace,amplitude=8pt},transform canvas={xshift=3.5pt}]
(m2-1-3.north east) -- (m2-9-3.south east) node (gSum) [midway,xshift=0.3cm,label=right:{\huge{$\Sigma$}}] {};
\node at (gSum.east) [right,xshift=3cm] (out) {output};
\draw (gSum.east) ++(1,0) edge [-stealth] node [above] {$f_\phi$} (out);
\end{tikzpicture}
	\caption{In the RN architecture pairs of objects $(o_i,o_j)$ (in red, green, blue) with input question q (yellow) are handled independently by the relation evaluation function $g_\theta$. The resulting vectors are summed up and the sum vector is transformed by $f_\phi$ to form the final network output.}
	\label{fig:RN}
\end{figure}

Relation Networks (RNs) as introduced by Santoro et al. (2017) \cite{santoro17} were a milestone in this direction. Their architecture is illustrated in Figure \ref{fig:RN}. RNs expect object vectors as input from previous layers or from samples, representing bits of information from the same domain. These objects can be as different as  outputs of recurrent layers encoding the information content of sentences in natural language processing or feature vectors representing abstract image features in the output of stacked convolutional layers. The objects’ relations to one another with respect to an input question $q$ is evaluated by an evaluation module $g$ for all combinatorial pairs of objects. The objects are expected to contain all information necessary to reason about their relation, most importantly also including their relative or absolute spatial or, in case of sequences, temporal position. The extracted relational information is then combined by summing the output vectors of $g$, which keeps the result invariant to the order of the objects and the order in which the pairs are evaluated. Finally, the combined vector is fed to an output function $f$, which leads to
\begin{equation}
	\label{eq:rn}
	RN(O,q)=f_\phi\Big(\sum_{i,j} g_\theta(o_i, o_j, q)\Big)
\end{equation}
as the final output of the RN. $f$ and $g$ are typically chosen as MLPs with weights $\phi$ and $\theta$, respectively. Since Relation Networks are differentiable if the internal functions are modeled accordingly, e.g. by using MLPs, the combination of object generating layers and Relation Network components can be trained end-to-end, so that suitable object representations can be learned inside the network. 

A number of extensions of the original RN have been introduced. In a recent paper by Palm et al. \cite{palm17} the network has one recurrent cell for each input object which is repeatedly fed its previous state, its input object, as well as transformed previous outputs from the other objects' recurrent cells. It can be shown that their Recurrent Relation Network is a special case of our approach, if our relation evaluation functions are made to share weights across all layers. However, doing so forces the inputs of each layer to reside in the same domain, thereby limiting layers' output representations to have a similar level of abstraction as the inputs. In another approach to enhance Relation Networks, Moon et al. \cite{moon18} proposed a memory enhanced variation with an attention mechanism which, while improving upon the RN results, doesn't solve all bAbI tasks. Other related architectures like RelNet by Bansal et al. \cite{bansal17} were benchmarked on each task separately, yet our approach achieves similar or better results despite being trained on all tasks jointly. 

We extend the RN simply by further layers, however, in a way that it scales linearly to arbitrary numbers of such layers. We stack these layers such that more complex relations can be infered, now being able to solve tasks that proved to be difficult in the past due to the number of related facts. The main advantage of our approach over other attempts to enhance RNs is its simplicity and its scalability, all while showing improved performance on the benchmark dataset.


\section{Multi-Layer Relation Networks}
The single-layer architecture discussed above only allows the relation evaluation function $g_\theta$ to consider information from two objects, making it unsuitable for problems requiring reasoning across more than two facts. While expanding $g$ to evaluate triplets would be possible, the computational complexity would increase from $O(n^2)$ to $O(n^3)$ with $n$ being the number of objects. Hence, simply scaling up the combinatorial layer is not feasable from a performance point of view. 

\begin{figure}
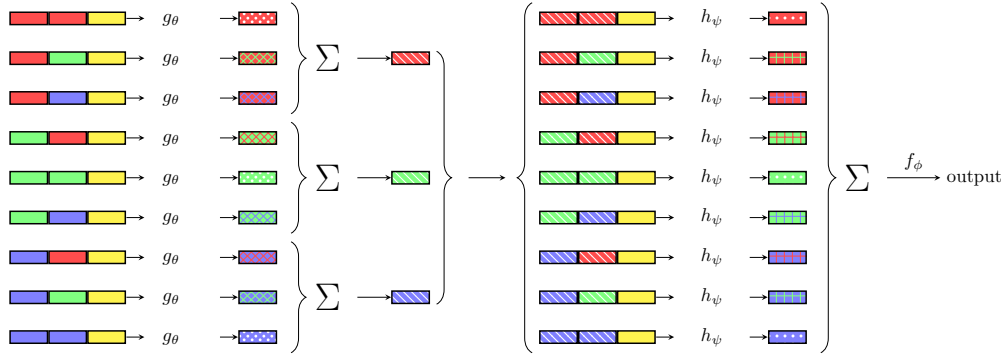

	\centering
	\includestandalone[scale=0.7]{dlrnTikz}
	\caption{Double-layer RN architecture: The output vectors of the relation evaluation function $g_\theta$ are grouped and summed by the first index $i$ of its input pairs $(o_i, o_j)$ (in red, green, blue) to form new object vectors (red/white, green/white, blue/white). The new objects are used as inputs for a single-layer RN (right half) with relation evaluation function $h_\psi$.}
	\label{fig:MLRN}
\end{figure}

Instead we sum up all the relations which reason about an object $o_i$, so that we end up with one sum vector per object. We then use these vectors as input objects for another single-layer Relation Network, thus forming a double-layer RN in total. The architecture of this Multi-Layer Relation-Network (MLRN) is shown in Figure \ref{fig:MLRN} and described by 
\begin{equation}
	DLRN(O,q)=f_\phi\bigg(\sum_{i,k} h_\psi\Big(\sum_j g_\theta(o_i, o_j, q), \sum_l g_\theta(o_k, o_l, q), q\Big)\bigg).
\end{equation}

This way the relation evaluation function $h_\psi$ reasons about objects which already contain binary relational knowledge. 
The construction process can be repeated for arbitrary layer stack counts $m$, with $m=0$ being equivalent to the single-layer RN. In a recursive formulation we obtain 
\begin{equation}
\label{eq:mlrn}
	MLRN(O,q)=f\Big(\sum_{i,k} h_{m, i, k}\Big)
\end{equation}
\begin{equation}
	h_{d, i, k}=h_d\left(\sum_j h_{d-1, i, j}, \sum_l h_{d-1, k, l}, q\right)\qquad d=1,\dots,m
\end{equation}
\begin{equation}
	h_{0, i, k}=g(o_i, o_k, q).
\end{equation}
Note that the MLRN has a computational complexity of $O(n^2 m)$, making it computationally suitable for more complex reasoning tasks.

\section{Experiments}
\subsection{The bAbI dataset}
Published by Weston et al. (2015) \cite{weston15}, the bAbI 20 QA tasks are a dataset of algorithmically generated sentences and related questions and answers formulated in natural language. Each task represents a type of reasoning as denoted in Table \ref{table:results}. The dataset contains 1000 or 10000 training/validation samples per task and 100 or 1000 test samples on the small and the large variant, respectively. For our work and all comparisons, we used the 10k variant with joint training on all tasks and predefined separation into 9k training samples, 1k validation samples and 1k test samples. A task is considered to be solved successfully if the test accuracy exceeds $95\%$. 

Under the same conditions, Differentiable Neural Computers (DNC) have managed to solve 18 of these 20 tasks \cite{santoro17}, Sparse DNCs 19/20 \cite{santoro17}, Relation Networks 18/20 \cite{santoro17}, Recurrent RNs 19/20 \cite{palm17}, Relation Memory Networks 19/20 \cite{moon18} and Working Memory Networks 19/20 \cite{pavez18}. While Adaptive Memory Networks \cite{li18} reported 20/20 successful tasks, their experiment settings mention different model training parameters for different types of tasks, hinting that the training was probably not performed on all tasks jointly. 

Examples of the most difficult tasks 2, 3, and 16, are given in the Appendix in Table \ref{table:examples}. Note that for task 16 about 6\% of the samples are not disambiguous as the context provides multiple possible answers and the answer provided in the label is not always the one with the most supporting facts.

\newcommand{\taskacc}[1]{
\begin{tikzpicture}
\begin{axis}[width=5.6cm,height=4.5cm,xlabel={training step},ylabel={accuracy},ymax=1.1,legend pos=south east,title={Task #1}]

\addplot[blue] table[x index=0,y index=1,col sep=comma] {merged_acc_qa#1.csv};
\addlegendentry{SLRN}

\addplot[red] table[x index=0,y index=2,col sep=comma] {merged_acc_qa#1.csv};
\addlegendentry{DLRN}

\end{axis}
\end{tikzpicture}
}

\subsection{Comparison of single- and double-layer RN}
Since choosing layer count m=0 yields an architecture which is equivalent to the original single-layer RN, we decided to reproduce the original results first and use the determined hyperparameters as a starting point for the multi-layer experiments.

As in the original RN paper we reduced or padded each sample’s context to contain exactly 20 sentences. While this preprocessing step might have an influence on the difficulty of the tasks since fewer unimportant sentences in the context equate to less input noise, preliminary tests of the MLRN indicated that this limitation does not have a significant impact on the network’s performance. To produce the input objects for the relation network, we use a word embedding matrix and feed the embedded word vectors, for each sentence separately, into an LSTM with peepholes. The final LSTM output is then concatenated with a position encoding vector which is a one-hot vector, encoding the position of the sentence inside the current sample’s context, with a randomized padding length per sample to reduce overfitting (as proposed by Adam Santoro, one of the authors of the original RN paper, in personal communication). The word embedding matrix and the LSTM weights were learned together with the RN weights during the end-to-end training.

\begin{table}[!]
\centering
\scalebox{0.8}{
\begin{tabular}{l | l l l}
	& Original RN results & Layer count $m=0$ & Layer count $m=1$ \\ \hline
	Task 1: Single Supporting Fact & $>95$ & $100$ & $100$ \\
	\bf Task 2: Two Supporting Facts & $\bf91.9$ & $\bf81.3$ & $\bf99.8$ \\
	\bf Task 3: Three Supporting Facts & $\bf83.5$ & $\bf90.5$ & $\bf99.3$ \\
	Task 4: Two Argument Relations & $>95$ & $100$ & $100$ \\
	Task 5: Three Argument Relations & $>95$ & $99.1$ & $99.9$ \\
	Task 6: Yes/No Questions & $>95$ & $100$ & $100$ \\
	Task 7: Counting & $>95$ & $100$ & $100$ \\
	Task 8: Lists/Sets & $>95$ & $100$ & $100$ \\
	Task 9: Simple Negation & $>95$ & $100$ & $100$ \\
	Task 10: Indefinite Knowledge & $>95$ & $99.4$ & $100$ \\
	Task 11: Basic Coreference & $>95$ & $99.1$ & $100$ \\
	Task 12: Conjunction & $>95$ & $100$ & $100$ \\
	Task 13: Compound Coreference & $>95$ & $100$ & $100$ \\
	Task 14: Time Reasoning & $>95$ & $99.9$ & $100$ \\
	Task 15: Basic Deduction & $>95$ & $100$ & $100$ \\
	Task 16: Basic Induction & $97.9$ & $97.0$ & $97.4$ \\
	Task 17: Positional Reasoning & $>95$ & $97.8$ & $98.3$ \\
	Task 18: Size Reasoning & $>95$ & $99.5$ & $99.6$ \\
	Task 19: Path Finding & $>95$ & $99.9$ & $100$ \\
	Task 20: Agent’s Motivations & $>95$ & $100$ & $100$ \\ \hline
	Tasks succeeded ($>95\%$) & $18/20$ & $18/20$ & $20/20$ \\
	Mean error & unknown & $1.825$ & $0.285$ \\
\end{tabular}
}
\caption{Task accuracies on the test set in \% for a single-layer RN as reported by the original authors, a single-layer RN with our generalized architecture ($m=0$) and our double-layer RN ($m=1$). Bold are the two tasks which significantly improved due to higher order relations.}
\label{table:results}
\end{table}

We chose a word embedding dimension of 256, an LSTM cell with 32 units, a position encoding vector size of 40, $f=MLP(256,512,dict\_size)$ with ReLU activations for the inner layers and a final linear layer with softmax for the output. For the relation evaluation function, we chose $g=MLP(256,256,256,256)$ with ReLU activations, added L2 weight regularization terms with a penalty of 2e-5 for the weights in f and g and used a cross-entropy loss function. For the gradient descent we used the adam optimizer with a fixed learning rate of 5e-5 and a batch size of 32. About 4 million training steps were necessary to finish the training, taking about 4 days on a single GTX 1080 Ti.


For the double-layer case ($m=1$), we used the same parameters as before, except switching to a weight penalty factor of 1e-4 and choosing $h_1=MLP(256,256,256)$. The training took 6 million steps, equating to about 8 days on a single GTX 1080 Ti.

For the sake of comparison tasks 2 and 3 are of most interest because they represent the failure cases of the single-layer RNs. Task 2 consists of questions with two supporting facts and task 3 represents questions with three supporting facts. 
Note that while our results for tasks 2 and 3 seem to be interchanged when compared to the results from the original paper, no evidence for a fault could be found. The results in Table \ref{table:results} show that the increased layer count does indeed allow the network to learn more complex relations. 

\begin{figure}
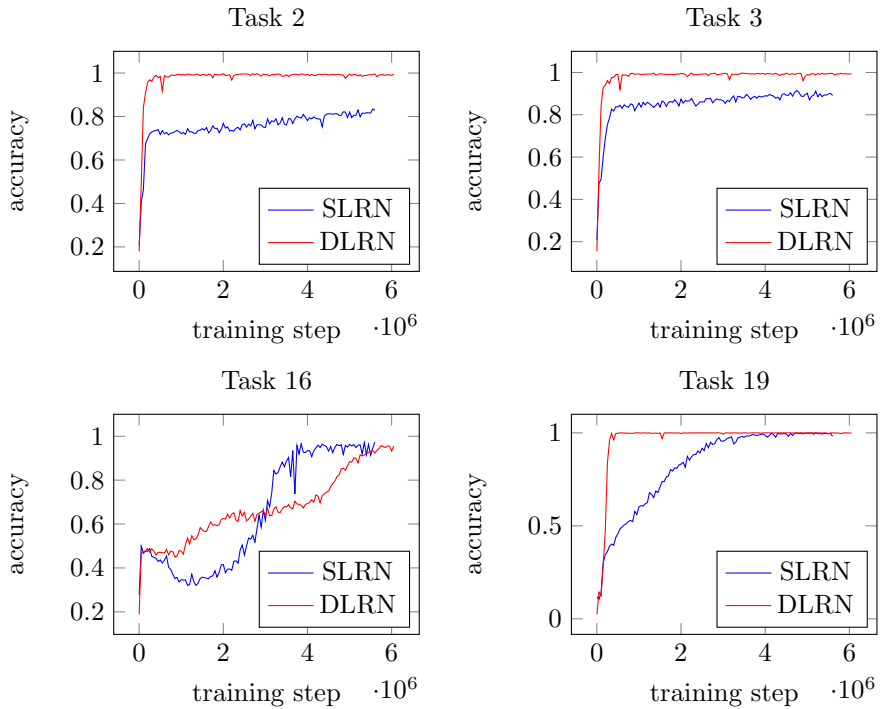

	\centering
	\begin{tabular}{cc}
\taskacc{2} & \taskacc{3} \\
\taskacc{16} & \taskacc{19} \\
	\end{tabular}
	\caption[Comparison of single layer RN and DLRN validation accuracy]{Accuracy on the validation set for a single layer RN and a double layer RN periodically measured during training.}
	\label{fig:accuracies}
\end{figure}

The MLRN not only learns more complex relations, but also learns more quickly. In Figure \ref{fig:accuracies} we show the learning curves on the validation sets of Tasks 2 and 3 and of Tasks 16 and 19.  As we can see, for reaching the smaller error in Task 2 and 3 no more training is necessary. The final error is reached very quickly. The faster learning is even more evident in Task 19. The single- and the double-layer RN both reach the same small error in this less complex task, but the MLRN reaches it with almost an order of magnitude less training steps. For Task 16 also our MLRN needs many training steps for learning, even more than the single-layer RN. Task 16 is the reason why the overall training on all tasks takes about twice as much time as for the single-layer RN. Also the single-layer RN has difficulties learning this task. We expect that these difficulties are due to the ambiguity of the training data, as shown in the Appendix. It is well known that the kind of symmetry breaking, which has to take place in such cases, is usually very time consuming. 

\section{Conclusion}
We have proposed a generalization of the Relation Network which can be scaled easily to multiple layers for higher task complexities. We have shown that this Multi-Layer Relation Network is capable of solving all of the bAbI QA tasks with joint training on all tasks, making it to our knowledge the first model to achieve this, all without resorting to using ensembles or taking the best of multiple runs. Its total test error rate of $0.285\%$ even matches the state-of-the-art of models trained on each task separately \cite{bansal17}. In most cases the MLRN also learns its tasks significantly faster than the single-layer RN, despite more parameters to train. This is a hint that its structure is better suited for these tasks. 

When compared to other proposed Relation Network enhancements, we attribute its advantage over Recurrent RNs \cite{palm17} to the fact that the shared weights in the recurrent architecture don't allow them to learn more complex relations in deeper layers because the weight-sharing forces all the layer outputs to be in the same domain, and therefore a similar level of abstraction, as the first layer's input.

With its simple construction the MLRN is suitable for a large variety of problems. In a next step we will evaluate our multi-layer architecture on further relational reasoning tasks like the relation of objects in images as in \cite{santoro17} or even challenges for abstract reasoning inspired by human IQ tests as proposed by \cite{barrett18}. \\

\noindent
Find our source code at \url{https://github.com/MMMMasterM/rn-variations}.

\newpage
\appendix
\setcounter{table}{0}
\renewcommand{\thetable}{A\arabic{table}}
\section{bAbI examples}
\begin{table}[h!]
\scalebox{0.80}{
\begin{tabular}{l | l}
	Task 2 & \texttt{\shortstack[l]{
				Context: \\
				1. Mary got the milk there. \\
				2. John moved to the bedroom. \\
				3. Sandra went back to the kitchen. \\
				4. Mary travelled to the hallway. \\
				\\
				Question: \\
				Where is the milk? \\
				\\
				Answer: \\
				hallway}} \\ \hline
	Task 3 & \texttt{\shortstack[l]{
				Context: \\
				1. John moved to the bedroom. \\
				2. John grabbed the apple there. \\
				3. Sandra moved to the hallway. \\
				4. John went to the office. \\
				5. Sandra went back to the bedroom. \\
				6. Sandra took the milk. \\
				7. John journeyed to the bathroom. \\
				8. John travelled to the office. \\
				9. Sandra left the milk. \\
				10. Mary went to the bedroom. \\
				11. Mary moved to the office. \\
				12. John travelled to the hallway. \\
				13. Sandra moved to the garden. \\
				14. Mary moved to the kitchen. \\
				15. Daniel took the football. \\
				16. Mary journeyed to the bedroom. \\
				17. Mary grabbed the milk there. \\
				18. Mary discarded the milk. \\
				19. John went to the garden. \\
				20. John discarded the apple there. \\
				\\
				Question: \\
				Where was the apple before the bathroom? \\
				\\
				Answer: \\
				office}} \\ \hline
	Task 16 (ambiguous) & \texttt{\shortstack[l]{
				Context: \\
				1. Greg is a frog. \\
				2. Bernhard is a swan. \\
				3. Julius is a frog. \\
				4. Bernhard is white. \\
				5. Julius is green. \\
				6. Lily is a frog. \\
				7. Brian is a frog. \\
				8. Lily is gray. \\
				9. Brian is gray. \\
				\\
				Question: \\
				What color is Greg? \\
				\\
				Answer: \\
				gray}} \\ \hline
	Task 16 (disambiguous) & \texttt{\shortstack[l]{
				Context: \\
				1. Lily is a swan. \\
				2. Bernhard is a lion. \\
				3. Greg is a swan. \\
				4. Bernhard is white. \\
				5. Brian is a lion. \\
				6. Lily is gray. \\
				7. Julius is a rhino. \\
				8. Julius is gray. \\
				9. Greg is gray. \\
				\\
				Question: \\
				What color is Brian? \\
				\\
				Answer: \\
				white}} \\
\end{tabular}
}
\caption{Samples from the most difficult tasks 2, 3 and 16 (one sample with and one without ambiguity).}
\label{table:examples}
\end{table}

\end{document}